%% file: main.tex
\documentclass[twoside,11pt]{article}

\input{command}

\begin{document}

\title{\thename: A Tabular Analytics and Learning Toolbox}

\author{\name Si-Yang Liu \email liusiyang@smail.nju.edu.cn \\
	\name Hao-Run Cai \email caihr@smail.nju.edu.cn \\
	\name Qi-Le Zhou \email zhouql@lamda.nju.edu.cn \\
	\name Han-Jia Ye \email yehj@lamda.nju.edu.cn\\
        \addr School of Artificial Intelligence, Nanjing University, China \\
        \addr National Key Laboratory for Novel Software Technology, Nanjing University, 210023, China\\
}
\maketitle

\input{abstract}

\input{introduction}

\input{mothod}
\input{toolbox}

\input{experiments}

\input{conclusion}

\bibliography{main}

\end{document}

%% file: command.tex
\usepackage{jmlr2e}
\usepackage{listings}
\usepackage{subfigure}
\usepackage{booktabs}
\usepackage{multirow}
\newcommand{\eg}{\emph{e.g.}}

\usepackage{tcolorbox}
\usepackage[noend]{algpseudocode}
\usepackage{xcolor,colortbl}
\definecolor{Gray}{gray}{0.85}
\definecolor{Redo}{rgb}{0.95,0.69,0.51}
\definecolor{LightCyan}{rgb}{0.88,1,1}
\usepackage{afterpage}
\usepackage{enumitem}
\usepackage{pifont}
\usepackage{tcolorbox}
\usepackage{verbatim}
\usepackage[misc,geometry]{ifsym}
\usepackage{tikz}
\usepackage{xspace}
\usetikzlibrary{shapes.geometric, arrows}

\newcommand{\thename}{{\sc Talent}\xspace}

\newcommand{\eat}[1]{}

\newcommand{\bbR}{\mathbb{R}}

\lstdefinestyle{lfonts}{
  basicstyle   = \footnotesize\ttfamily,
  stringstyle  = \color{purple},
  keywordstyle = \color{blue!90!black}\bfseries,
  commentstyle = \color{green!60!black},
}
\lstdefinestyle{lnumbers}{
  numbers     = left,
  numberstyle = \tiny,
  numbersep   = 1em,
  firstnumber = 1,
  stepnumber  = 1,
}
\lstdefinestyle{llayout}{
  breaklines       = true,
  tabsize          = 2,
  columns          = flexible,
}
\lstdefinestyle{lgeometry}{
  xleftmargin      = 20pt,
  xrightmargin     = 0pt,
  frame            = tb,
  framesep         = \fboxsep,
  framexleftmargin = 20pt,
}
\lstdefinestyle{lgeneral}{
  style = lfonts,
  style = lnumbers,
  style = llayout,
  style = lgeometry,
}
\lstdefinestyle{python}{
    language = {Python},
    style    = lgeneral,
}

\ShortHeadings{Technical report}{Technical report}
\firstpageno{1}

%% file: abstract.tex
\begin{abstract}
Tabular data is one of the most common data sources in machine learning. Although a wide range of classical methods demonstrate practical utilities in this field, deep learning methods on tabular data are becoming promising alternatives due to their flexibility and ability to capture complex interactions within the data. Considering that deep tabular methods have diverse design philosophies, including the ways they handle features, design learning objectives, and construct model architectures, we introduce a versatile deep-learning toolbox called \thename (\textbf{T}abular \textbf{A}nalytics and \textbf{LE}ar\textbf{N}ing \textbf{T}oolbox) to utilize, analyze, and compare tabular methods.
\thename encompasses an extensive collection of more than 20 deep tabular prediction methods, associated with various encoding and normalization modules, and provides a unified interface that is easily integrable with new methods as they emerge. In this paper, we present the design and functionality of the toolbox, illustrate its practical application through several case studies, and investigate the performance of various methods fairly based on our toolbox.
Code is available at~\href{https://github.com/qile2000/LAMDA-TALENT}{https://github.com/qile2000/LAMDA-TALENT}.
\end{abstract}

\begin{keywords}
Tabular Data, Deep Learning, Deep Tabular Prediction, Machine Learning
\end{keywords}

%% file: introduction.tex
\section{Introduction}
Machine learning has achieved remarkable success across a broad spectrum of domains. Tabular data, characterized by datasets arranged in a table format, represents one of the most prevalent types of data used in machine learning applications such as click-through rate (CTR) prediction~\citep{DeepFMCTR}, cybersecurity~\citep{cybersecurity}, medical analysis~\citep{schwartz2007drug}, and identity protection~\citep{identityprotection2022}. In these datasets, each row typically represents an individual instance, while each column corresponds to a different attribute or feature. In the context of supervised learning, each training instance is associated with a label, which can be discrete for classification tasks or continuous for regression tasks as shown in~\autoref{tab:tabular_example}. Machine learning models are designed to learn a mapping from instances to their labels using the training data, with the aim to generalize this mapping to unseen instances from the same distribution.

\begin{table}[h]
\centering
\begin{tcolorbox}[colback=gray!10, boxrule=0.5pt, arc=4pt, boxsep=0pt, left=5pt, right=5pt, top=6pt, bottom=6pt]
    \begin{tabular}{@{}ccccccc@{}}
        \toprule
        \textbf{Age} & \textbf{Education} & \textbf{Occupation} &\textbf{Race} & \textbf{Sex} &\textbf{hours-per-week} & \textbf{Income} \\
        \midrule
        39 & Bachelors & Adm-clerical &White & Male &40 & $\leq$50K \\
        50 & Bachelors & Exec-managerial &White & Male & 13 & $>$50K \\
        38 & HS-grad & Handlers-cleaners &White & Male &40 & $\leq$50K \\
        53 & 11th & Handlers-cleaners &Black & Male & 40& $\leq$50K \\
        28 & Bachelors & Prof-specialty &Black & Female &40 & $>$50K \\
        \midrule
        45 & Masters & Exec-managerial & White  & Female & 50 & - \\
        \bottomrule
    \end{tabular}
\end{tcolorbox}
\caption{An example of a binary classification task from the Adult dataset~\citep{misc_adult_2}. The first six attributes/features (columns) are used to predict the final label. The first five rows are training examples, and the last row is a test instance with an unknown income.}
\label{tab:tabular_example}
\end{table}


Methodologies for analyzing tabular datasets have significantly evolved. Classical techniques such as Logistic Regression (LogReg), Support Vector Machine (SVM), Multi-Layer Perceptron (MLP), and decision tree have long served as the foundation of numerous algorithms~\citep{bishop2006pattern}. In practical applications, tree-based ensemble methods like XGBoost~\citep{chen2016xgboost}, LightGBM~\citep{ke2017lightgbm}, and CatBoost~\citep{Prokhorenkova2018Catboost} have demonstrated substantial improvements in performance. Inspired by the achievements of Deep Neural Networks (DNNs) in visual and linguistic tasks~\citep{simonyan2014very,vaswani2017attention,devlin2018bert}, researchers have recently developed deep learning models specifically for tabular data~\citep{ZhangDW16Deep,Borisov2022Deep}. 
While initial deep learning approaches for tabular data encountered challenges due to their inherent complexity, ongoing advancements have increasingly focused on enhancing complex feature interaction modeling and mimicking the decision-making processes found in tree-based models~\citep{Cheng2016Wide,DeepFMCTR,PopovMB20Neural,Chang0G22NODEGAM}. 
Continuous research has shown that modern deep learning techniques can dramatically improve upon the performance of traditional models such as MLPs~\citep{ArikP21TabNet,GorishniyRKB21Revisiting,Kadra2021Well}. These advanced deep tabular models are able to effectively model complex relationships among instances or features, uncover underlying patterns in the datasets, and improve prediction performance.

While deep learning offers significant benefits for analyzing tabular data, its practical application is often hindered by the lack of uniform interfaces and varying preprocessing demands among different methods. We introduce a versatile and powerful toolbox, TALENT (\textbf{T}abular \textbf{A}nalytics and \textbf{LE}ar\textbf{N}ing \textbf{T}oolbox), for tabular data prediction.
\thename integrates diverse methodologies, including classical methods as well as advanced deep methods, into a unifying framework. \thename not only standardizes interfaces and streamlines preprocessing steps, making it easy to integrate new methods as they emerge, but also ensures that all methods can be fairly compared, providing a reliable basis for evaluating their effectiveness in different scenarios.
More importantly, our toolbox enables the composition of effective deep tabular modules and facilitates data analysis, offering scalable solutions that can adapt to various complexities and data-specific needs.
\clearpage
The advantages of the toolbox are
\begin{itemize}
\item \textbf{Model Diversity}. Our toolbox integrates an extensive array of more than 20 diverse deep tabular methods with uniform interfaces, allowing users to select the best-fit model based on the complexity and specifics of their tasks.
\item \textbf{Encoding Techniques}. In addition to various encoding strategies for categorical features, \thename provides eight encoding techniques for numerical features. These comprehensive encoding techniques ensure versatile data representation tailored to different analytical requirements.
\item \textbf{Extensibility}. 	The modular architecture of the toolbox ensures flexibility and future scalability. Users can easily add new models and methods according to practical requirements, making it a continuously relevant and valuable resource.
\end{itemize}

%% file: mothod.tex
\section{\thename for Tabular Prediction}

We formally define the tabular prediction task and then provide an overview of the tabular prediction
methods supported by {\thename}.

\subsection{Preliminary}
A supervised tabular dataset is formatted as $N$ examples and $d$ features, corresponding to $N$ rows and $d$ columns in the table. An instance $x_i\in \mathbb R {^d}$ is depicted by its $d$ feature values. 
Assume $x_{i,j}$ as the $j$-th feature of instance $x_i$, it could be a numerical (continuous) one $x_{i,j}^{\textit{\rm num}}\in\bbR$, or a categorical (discrete) value $x_{i,j}^{\textit{\rm cat}}$. 
The categorical features are usually transformed in an index (integer).
Each instance is associated with a label $y_i$, where $y_i\in \{1,-1\}$ in a binary classification task, $y_i\in [C]=\{1,\ldots,C\}$ in a multi-class classification task, and $y_i\in \bbR$ in a regression task. 
Given a tabular dataset $\mathcal D=\{(x_i, y_i)\}_{i=1}^N$, we aim to learn a model $f$ on $\mathcal D$ via empirical risk minimization that maps $x_i$ to its label $y_i$:
\begin{equation}
\min_f \; \sum_{(x_i, y_i)\in\mathcal D}  \ell(y, \;\hat{y}_i=f(x_i)) + \Omega(f)\;.\label{eq:objective}
\end{equation}
$\ell(\cdot, \cdot)$ measures the discrepancy between the predicted label $\hat{y}_i$ and the true label $y_i$, \eg, cross-entropy in classification. 
$\Omega(\cdot)$ is the regularization on the model.
We expect the learned $f$ is able to extend its ability to unseen instances sampled from the same distribution as $\mathcal D$. \looseness=-1

\subsection{Supported Methods}
In \thename, we implement a comprehensive range of models implement the mapping \( f \) from features to outputs with the same interface, covering classical methods, tree-based methods, and advanced deep tabular methods. 

Classical models in \thename include K-Nearest Neighbors (KNN) and SVM for various tasks, complemented by Linear Regression (LR) for regression tasks and  Logistic Regression (LogReg), Naive Bayes, and Nearest Class Mean (NCM) for classification.

Tree-based methods in \thename utilize powerful algorithms, including Random Forest, XGBoost~\citep{chen2016xgboost}, CatBoost~\citep{Prokhorenkova2018Catboost} and LightGBM~\citep{ke2017lightgbm}, known for their high efficiency and strong predictive performance across a variety of datasets.

\begin{figure}[t]
  \centering
  \includegraphics[width=\textwidth]{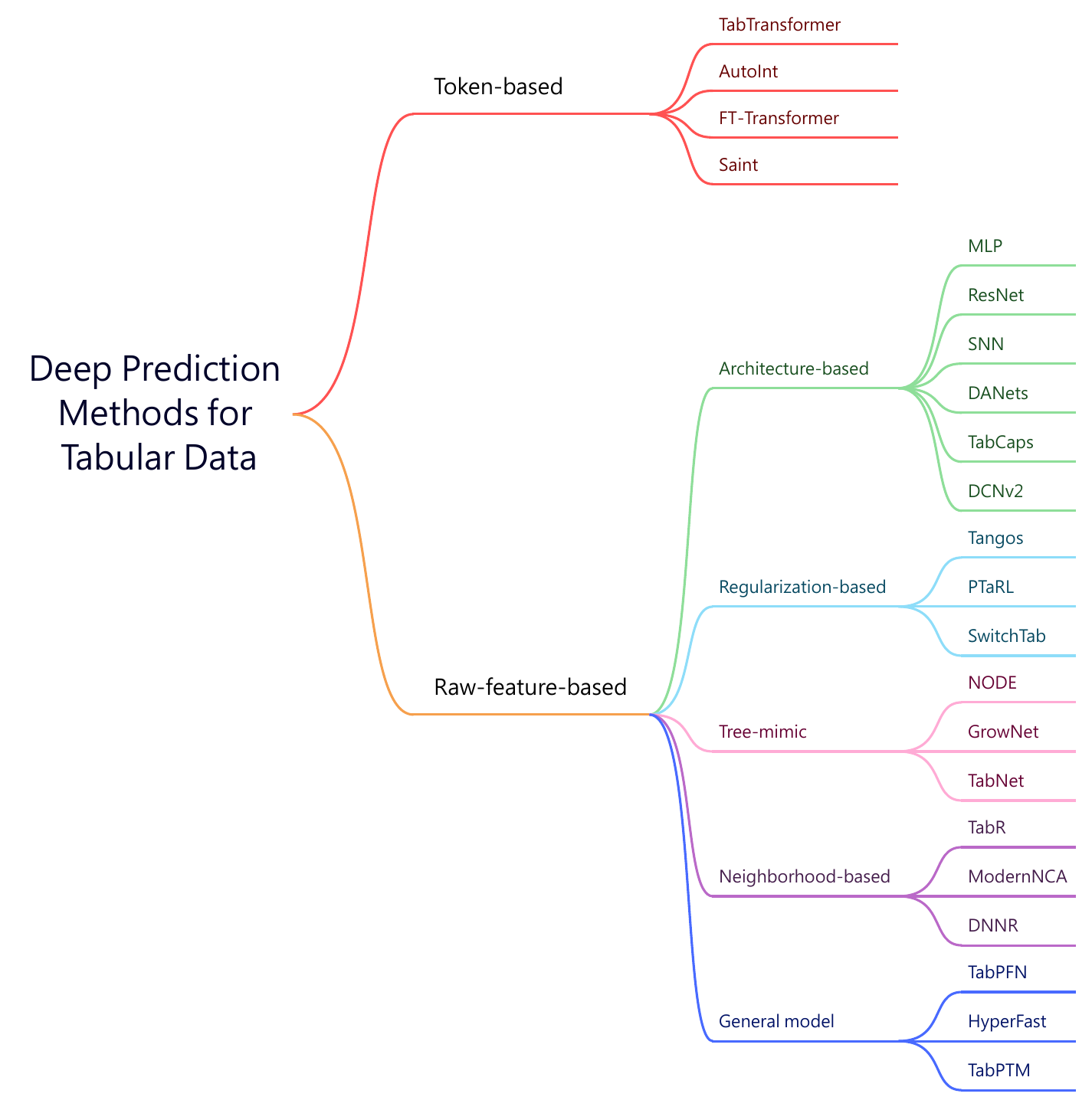}
  \caption{Various deep prediction methods for tabular data in \thename. }
  \label{fig:methods_tree}
\end{figure}
Our toolbox provides a comprehensive selection of representative deep tabular prediction methods, each meticulously designed to tackle distinct challenges in tabular data analysis. A taxonomy of these methods is shown in~\autoref{fig:methods_tree}:
\begin{itemize}[noitemsep,topsep=0pt,leftmargin=*]
    \item \textbf{MLP}: A multi-layer neural network, implemented according to~\cite{GorishniyRKB21Revisiting}.
    \item \textbf{ResNet}: A DNN that uses skip connections across many layers, which is implemented according to~\cite{GorishniyRKB21Revisiting}.
    \item \textbf{SNN}~\citep{KlambauerUMH17Self}: An MLP-like architecture utilizing the SELU activation, which facilitates the training of deeper neural networks.
    \item \textbf{DANets}~\citep{ChenLWCW22DAN}: A neural network designed to enhance tabular data processing by grouping correlated features and reducing computational complexity.
    \item \textbf{TabCaps}~\citep{Chen2023TabCaps}: A capsule network that encapsulates all feature values of a record into vectorial features.
    \item \textbf{DCNv2}~\citep{WangSCJLHC21DCNv2} consists of an MLP-like module combined with a feature crossing module, which includes both linear layers and multiplications.
    \item \textbf{NODE}~\citep{PopovMB20Neural}: A tree-mimic method that generalizes oblivious decision trees, combining gradient-based optimization with hierarchical representation learning.    
    \item \textbf{GrowNet}~\citep{Badirli2020GrowNet}: A gradient boosting framework that uses shallow neural networks as weak learners. 
    \item \textbf{TabNet}~\citep{ArikP21TabNet}: A tree-mimic method using sequential attention for feature selection, offering interpretability and self-supervised learning capabilities.
    \item \textbf{TabR}~\citep{gorishniy2023tabr}: A deep learning model that integrates a KNN component to enhance tabular data predictions through an efficient attention-like mechanism.
    \item \textbf{ModernNCA}~\citep{ModernNCA2024}: A deep tabular model inspired by traditional Neighbor Component Analysis~\citep{NCA2004}, which makes predictions based on the relationships with neighbors in a learned embedding space.
    \item \textbf{DNNR}~\citep{NaderSL22DNNR} enhances KNN by using local gradients and Taylor approximations for more accurate and interpretable predictions.
    \item \textbf{AutoInt}~\citep{SongS0DX0T19AutoInt}: A token-based method that uses a multi-head self-attentive neural network to automatically learn high-order feature interactions.
    \item \textbf{Saint}~\citep{Somepalli2021SAINT}: A token-based method that leverages row and column attention mechanisms for tabular data.
    \item \textbf{TabTransformer}~\citep{Huang2020TabTransformer}: A token-based method that enhances tabular data modeling by transforming categorical features into contextual embeddings.
    \item \textbf{FT-Transformer}~\citep{GorishniyRKB21Revisiting}: A token-based method which transforms features to embeddings and applies a series of attention-based transformations to the embeddings.
    \item \textbf{TANGOS}~\citep{jeffares2023tangos}: A regularization-based method for tabular data that uses gradient attributions to encourage neuron specialization and orthogonalization.
    \item \textbf{SwitchTab}~\citep{Wu2024SwitchTab}: A self-supervised method tailored for tabular data that improves representation learning through an asymmetric encoder-decoder framework.
    \item \textbf{PTaRL}~\citep{PTARL}: A regularization-based framework that enhances prediction by constructing and projecting into a prototype-based space. \looseness=-1
    \item \textbf{TabPFN}~\citep{Hollmann2022TabPFN}: A general model which involves the use of pre-trained deep neural networks that can be directly applied to other tabular classification tasks.
    \item \textbf{HyperFast}~\citep{HyperFast2024}: A meta-trained hypernetwork that generates task-specific neural networks for instant classification of tabular data.
    \item \textbf{TabPTM}~\citep{TabPTM2023}: A general method for tabular data that standardizes heterogeneous datasets using meta-representations, allowing a pre-trained model to generalize to unseen datasets without additional training.
\end{itemize}

\subsection{Encoding Techniques}
According to~\cite{Numerical_embedding}, embeddings for numerical features greatly improve the performance of deep learning models on tabular data by providing more expressive and powerful initial representations. This approach is useful for both MLPs and advanced Transformer-like architectures. In \thename, we incorporate various numerical encoding techniques, enhancing the input quality for machine learning models. The diverse range of encoding methods ensures effective and customized data preprocessing for different analytical needs. Here are the encoding methods included in \thename:
\begin{itemize}[noitemsep,topsep=0pt,leftmargin=*]
    \item \textbf{Quantile-based Binning (Q\_bins)} constructs bins by dividing value ranges according to the quantiles of the individual feature distributions, and replaces the original values with their corresponding bin indices.

    \item \textbf{Target-aware Binning (T\_bins)} creates bins using training labels to correspond to narrow ranges of possible target values. This approach is similar to the ``C4.5 Discretization'' algorithm~\citep{C451996}, which splits the value range of each feature using the target as guidance.

    \item \textbf{Quantile-based Unary Encoding (Q\_Unary)}~\citep{Unary} converts numerical values into unary binary-encoded bin indices based on quantiles.

    \item \textbf{Target-aware Unary Encoding (T\_Unary)}~\citep{Unary} generates unary binary-encoded bin indices using target-aware transformations.

    \item \textbf{Quantile-based Johnson Encoding (Q\_Johnson)}~\citep{Label_encoding_for_regression} encodes numerical data based on quantile intervals using Johnson distribution transformations~\citep{JohnsonEncoding}, replacing original values with Johnson binary-encoded bin indices.

    \item \textbf{Target-aware Johnson Encoding (T\_Johnson)}~\citep{Label_encoding_for_regression} applies Johnson transformations with target-aware bins, replacing original values with Johnson binary-encoded bin indices~\citep{JohnsonEncoding}.

    \item \textbf{Quantile-based Piecewise Linear Encoding (Q\_PLE)}~\citep{Numerical_embedding} segments numerical data based on quantiles and applies piecewise linear transformations.

    \item \textbf{Target-aware Piecewise Linear Encoding (T\_PLE)}~\citep{Numerical_embedding} builds target-aware bins and applies piecewise linear transformations.
\end{itemize}

Additionally, \thename employs various categorical encoding techniques, including \textbf{Ordinal} encoding, \textbf{One-Hot} encoding, \textbf{Binary} encoding, \textbf{Hash} encoding~\citep{HashEncoder}, \textbf{Target} encoding~\citep{TargetEncoder}, \textbf{Leave-One-Out} encoding, and \textbf{CatBoost} encoding~\citep{Prokhorenkova2018Catboost}.

%% file: toolbox.tex
\section{Toolbox Usage}
In this section, we introduce the dependencies and workflow when using \thename. 

\subsection{Dependencies}

\thename leverages open-source libraries to support its advanced data processing and machine learning functionalities, following the organized code structure introduced in rtdl~\citep{GorishniyRKB21Revisiting}. For model optimization and hyperparameter tuning, it utilizes Optuna~\citep{akiba2019optuna}. These dependencies are carefully selected, providing users with a powerful, flexible, and efficient toolbox for tackling diverse challenges in the analysis of tabular data.

\subsection{The workflow of \thename}
\begin{figure}[h]
  \centering
  \includegraphics[width=\textwidth]{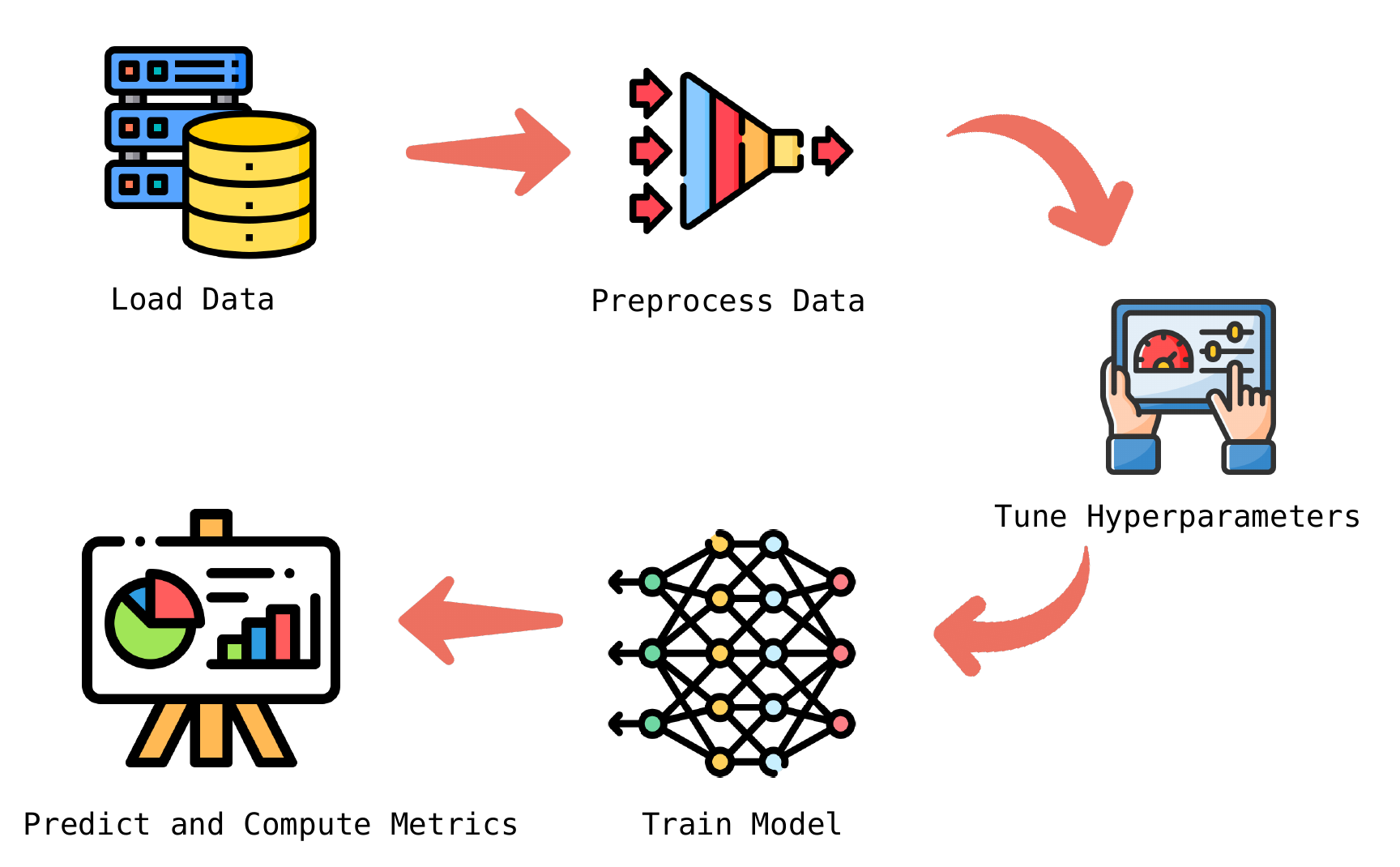}
  \caption{Flowchart depicting the data prediction process with {\sc{Talent}}.}
  \label{fig:TALENT_flowchart}
\end{figure}

 The flowchart in~\autoref{fig:TALENT_flowchart} visually represents the streamlined workflow facilitated by our toolbox. It begins with loading the data, followed by preprocessing,  hyperparameter tuning, model training, prediction, and finally, evaluation. This structured process ensures a smooth transition from raw data to meaningful results.

\thename offers interfaces to over 30 methods as mentioned above.  When using \thename, users have the flexibility to configure hyperparameters within the JSON files stored in the \texttt{configs} folder. Users can either modify the hyperparameters directly in the JSON files or override them through command-line inputs.

For example, to run a classical method, \eg, \texttt{xgboost},  using the default hyperparameters from the JSON configuration file, users would use the following command:

\begin{lstlisting}[language=sh, backgroundcolor=\color{LightCyan}]
    python train_model_classical.py --model_type xgboost
\end{lstlisting}

To run a deep learning method, \eg, \texttt{mlp}, use the following command:

\begin{lstlisting}[language=sh, backgroundcolor=\color{LightCyan}]
    python train_model_deep.py --model_type mlp
\end{lstlisting}

The command-line arguments for running the models are described below, allowing for customization depending on the specific requirements of the task and the dataset:

\begin{itemize}[noitemsep,topsep=0pt,leftmargin=*]
    \item \textbf{model\_type} specifies the model to be used.
    \item \textbf{dataset} defines the dataset to be used, specified by the name of the dataset folder.
    \item \textbf{max\_epoch} sets the maximum number of epochs for training the model.
    \item \textbf{batch\_size} determines the size of the batch of samples to be processed per gradient update.
    \item \textbf{seed\_num} specifies the number of different seed values to use for the experiments, ensuring multiple runs for statistical robustness.
    \item \textbf{normalization} chooses the type of normalization to apply to the dataset, including \textbf{Standard} scaling, \textbf{MinMax} scaling, \textbf{Quantile transformation}, \textbf{MaxAbs} scaling, \textbf{Power transformation}, and \textbf{Robust} scaling.
    \item \textbf{num\_nan\_policy:} Defines the policy for handling missing numerical data, such as \textbf{mean, median}.
    \item \textbf{cat\_nan\_policy} specifies the policy for handling missing categorical data, such as \textbf{most\_frequent, constant}.
    \item \textbf{cat\_policy} determines the categorical encoding method to be applied.
     \item \textbf{num\_policy} determines the numerical encoding method to be applied.
    \item \textbf{n\_trials} sets the number of trials for hyperparameter tuning (if applicable).
    \item \textbf{tune} indicates whether hyperparameter tuning should be performed, with options typically being \textbf{True} or \textbf{False}.
\end{itemize}

By using \thename, users can benefit from its configuration and evaluation interface, ensuring that experiments are both reproducible and customizable to meet specific research needs.

\subsection{Model Default Hyperparameters and Search Space}

For each model supported by \thename, there are two essential JSON files that aid in the configuration and search of  hyperparameters of the model:

\begin{itemize}[noitemsep,topsep=0pt,leftmargin=*]
    \item \textbf{Default Hyperparameters}: Each model has a corresponding JSON file in the \\ \texttt{configs/default} folder containing the default hyperparameters. These hyperparameters are pre-defined based on either values used in the literature or empirically determined settings that provide good performance.

    \item \textbf{Hyperparameter Search Space}: Each model also has a corresponding JSON file in the \texttt{configs/opt\_space} folder that defines the hyperparameter search space for the tuning process. This file specifies the bounds and types of each hyperparameter (\eg, continuous, discrete, categorical) that can be explored using automated hyperparameter optimization tools like Optuna~\citep{akiba2019optuna}. This setup facilitates a more systematic and potentially more effective search for optimal model configurations, especially useful when adapting models to new datasets or specific tasks.
\end{itemize}

These configuration files ensure that users not only have a reliable starting point for each model but also the flexibility to explore and optimize hyperparameters to match specific data characteristics or objectives.

For instance, when the users choose to use MLP, we have:

\begin{itemize}[noitemsep,topsep=0pt,leftmargin=*]
    \item \textbf{configs/default/mlp.json}:
\begin{lstlisting}[language=python,backgroundcolor=\color{LightCyan}]
{
    "mlp": {
        "model": {
            "d_layers": [384, 384], 
            "dropout": 0.1
        },
        "training": {
            "lr": 3e-4,
            "weight_decay": 1e-5
        }
    }
}
\end{lstlisting}

    \item \textbf{configs/opt\_space/mlp.json}:
\begin{lstlisting}[language=python,backgroundcolor=\color{LightCyan}]
{
    "mlp": {
        "model": {
            "d_layers": ["$mlp_d_layers", 1, 8, 64, 512],
            "dropout": ["?uniform", 0.0, 0.0, 0.5]
        },
        "training": {
            "lr": ["loguniform", 1e-05, 0.01],
            "weight_decay": ["?loguniform", 0.0, 1e-06, 0.001]
        }
    }
}
\end{lstlisting}
\end{itemize}

\subsection{Example Usage}

Below is an example of how to use the toolbox to run experiments across different seeds, allowing for robust evaluation of the performance of methods:

\begin{lstlisting}[style=python, morekeywords={evaluate}]
    # Parse the arguments and load default parameters and optimization space
    args, default_para, opt_space = get_args()
    # Load the training, validation, and test data
    train_val_data, test_data, info = get_dataset(args.dataset, args.dataset_path)
    # If hyperparameter tuning is enabled, tune the hyperparameters
    if args.tune:
        args = tune_hyper_parameters(args, opt_space, train_val_data, info)
    ## Training stage over different random seeds
    for seed in tqdm(range(args.seed_num)):
        args.seed = seed  # Update seed for reproducibility
        # Get the method based on the model type
        method = get_method(args.model_type)(args, info['task_type'] == 'regression')
        # Train the model and record the time cost
        time_cost = method.fit(train_val_data, info, train=True)
        # Predict using the trained model on the test data 
        # and calculate various metrics
        vres, metric_name, predict_logits = method.predict(test_data, info)
\end{lstlisting}

In this script:
\begin{itemize}[noitemsep,topsep=0pt,leftmargin=*]
    \item The \textbf{get\_args} function retrieves and parses the default arguments, default hyperparameters, and the optimization space for hyperparameter tuning.
    \item The \textbf{get\_dataset} function loads the specified dataset from the given path, splits it into training/validation and test sets, and provides additional information about the dataset.
    \item If hyperparameter tuning is enabled, the \textbf{tune\_hyper\_parameters} function adjusts the arguments based on the optimization space and the training/validation data.
    \item The \textbf{get\_method} function selects the appropriate modeling class based on the model type specified in \textbf{args.model\_type}.
    \item The seed is updated for each iteration, ensuring that each run is reproducible but distinct, enhancing the statistical robustness of the results.
    \item The performance metrics and predictions are recorded for each seed, allowing for a comprehensive evaluation of the model across different initializations. For classification tasks, the evaluation metrics include Accuracy, Average Recall, Average Precision, F1 Score, LogLoss, and AUC. For regression tasks, the metrics include MAE, RMSE, and R2~\citep{lewis2015applied}.
\end{itemize}

\subsection{Adding New Methods}

\begin{figure}[h]
  \centering
  \includegraphics[width=\textwidth]{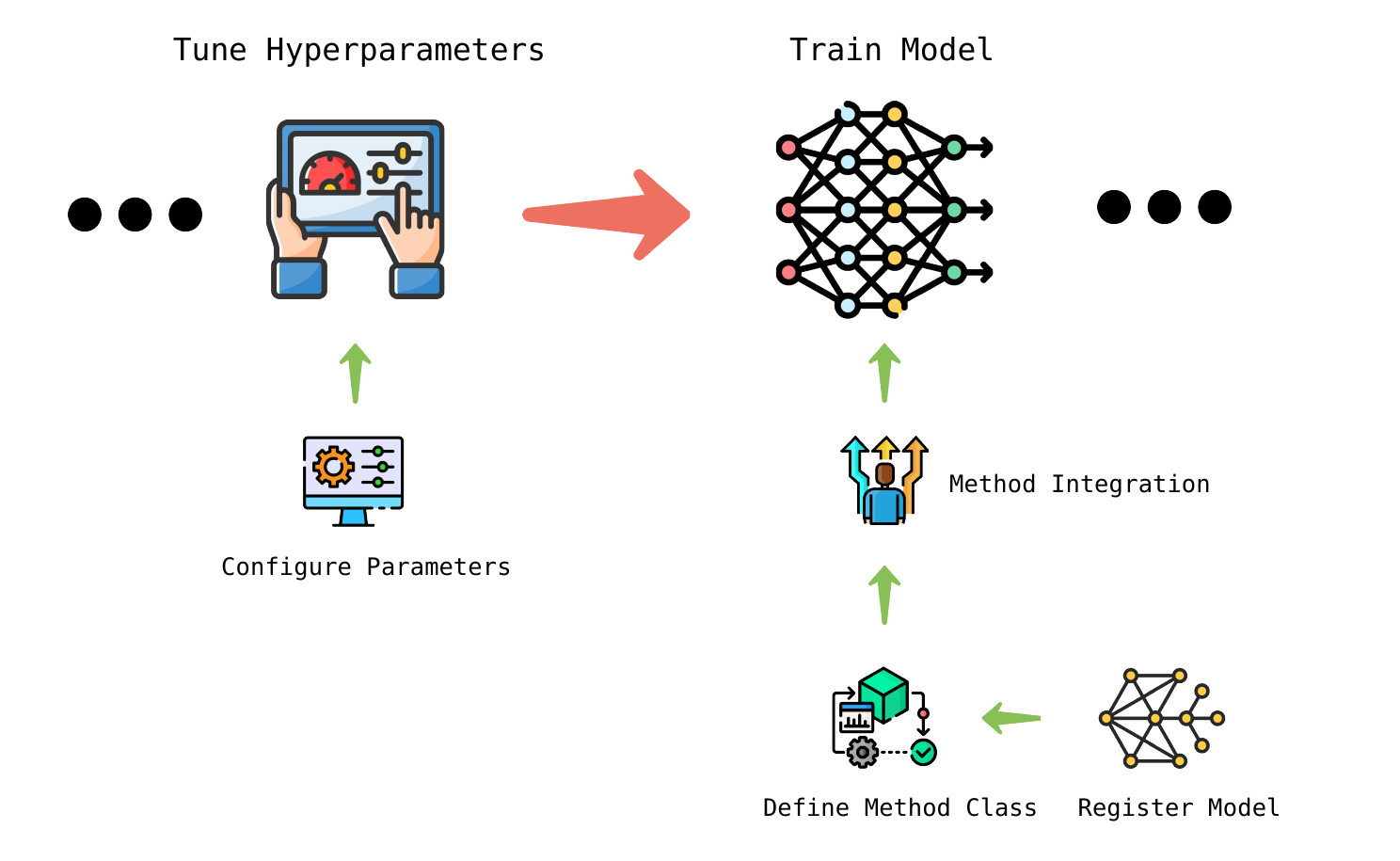}
  \caption{Workflow for Adding a New Method to \thename.}
  \label{fig:add_method_workflow}
\end{figure}
\thename is designed to be highly customizable, allowing users to integrate new machine learning methods effortlessly. Whether users are adding an well-known algorithm or experimenting with a novel approach, follow these steps to expand the capabilities of the toolbox, as illustrated in~\autoref{fig:add_method_workflow}:

\begin{enumerate}[noitemsep,topsep=0pt,leftmargin=*]
    \item \textbf{Register the Model:} Start by registering the new model class in the \texttt{model/models} directory. Ensure that this class includes the architecture of the model, defining how the model will be constructed.

    \item \textbf{Create the Method Class:} Create a new method class within the \texttt{model/methods} directory. This class should inherit from the base class provided in \texttt{base.py}. Implement the necessary components of the machine learning method in this class, including the training and prediction processes.

    \item \textbf{Method Integration:} Integrate the new method into the workflow of \thename by adding its name to the \texttt{get\_method} function located in \texttt{model/utils.py}. This function maps model types to their respective classes, enabling the toolbox to instantiate the correct model.

    \item \textbf{Configure Parameters:} Update the JSON files in the \texttt{configs/default} and \\ \texttt{configs/opt\_space} directories to include default hyperparameters and hyperparameter search spaces for the new method. 

    \item \textbf{Adjust Training Processes:} If the method requires a unique training procedure, modify the relevant functions in \texttt{model/methods/base.py}. Tailor these functions to accommodate any special optimization strategies that the method requires.
\end{enumerate}

By following these steps, researchers can add new algorithms to \thename, adapting it to meet diverse research needs. For detailed examples and additional guidance, refer to the implementation of existing methods in the \texttt{model/methods} directory.

%% file: experiments.tex
\section{Preliminary Experiments}
\begin{figure}[h]
  \centering
   \begin{minipage}{0.43\linewidth}
    \includegraphics[width=\textwidth]{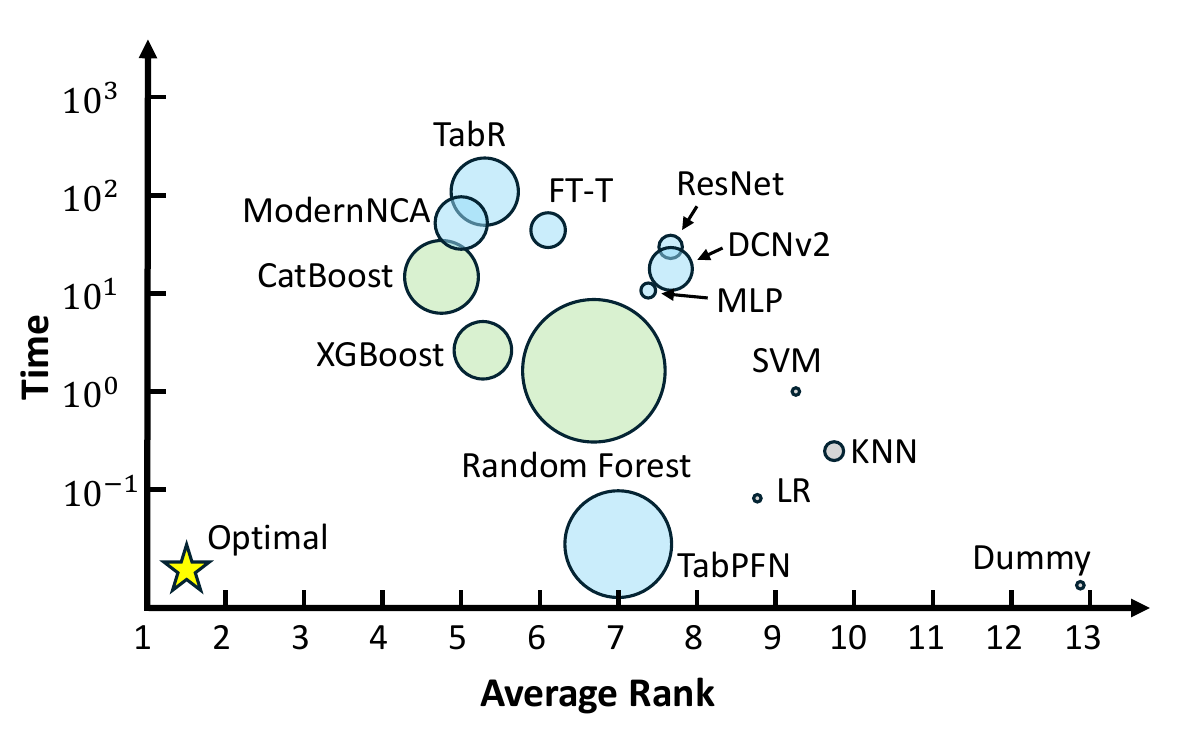}
    \centering
    {\small \mbox{(a) {Binary Classification}}}
    \end{minipage}
    \begin{minipage}{0.43\linewidth}
    \includegraphics[width=\textwidth]{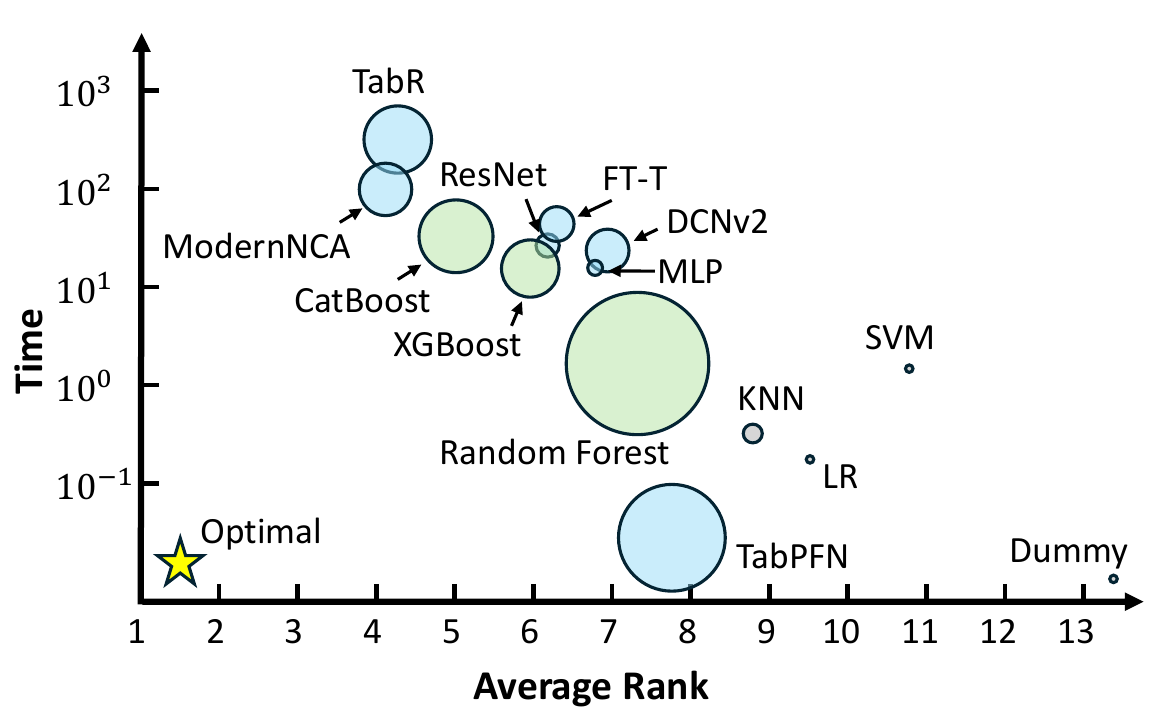}
    \centering
    {\small \mbox{(b) {Multi-Class Classification}}}
    \end{minipage}
    \\
    \begin{minipage}{0.43\linewidth}
    \includegraphics[width=\textwidth]{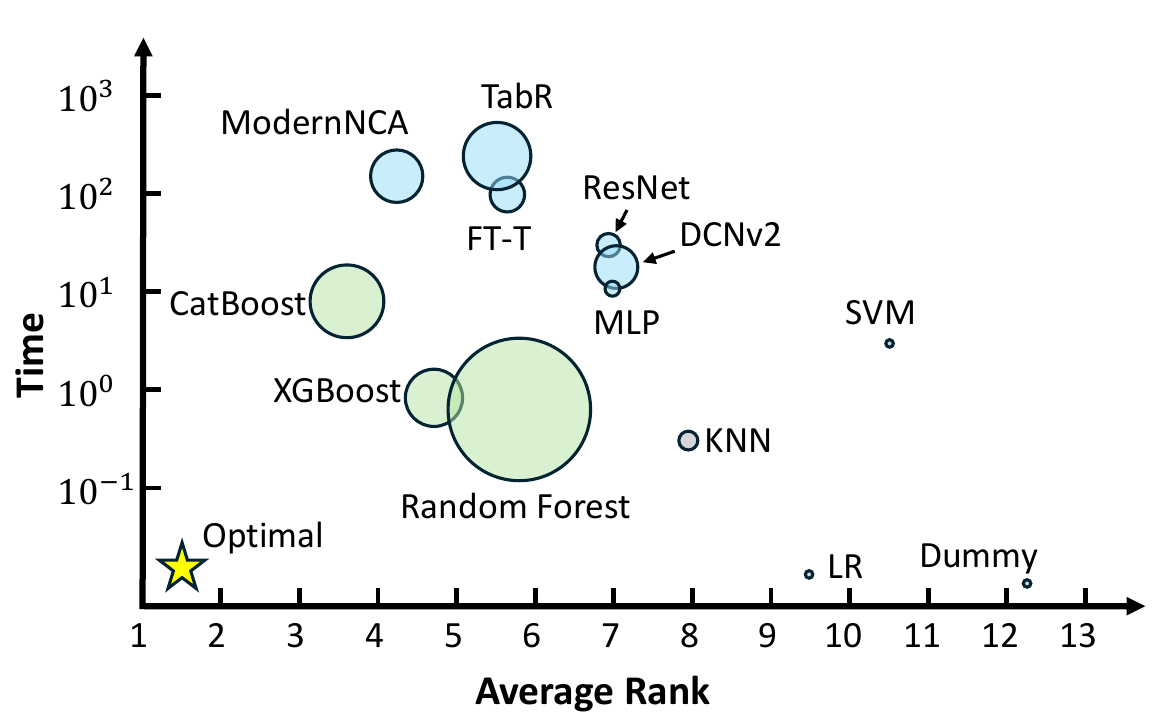}
    \centering
    {\small \mbox{(c) {Regression}}}
    \end{minipage}
    \begin{minipage}{0.43\linewidth}
    \includegraphics[width=\textwidth]{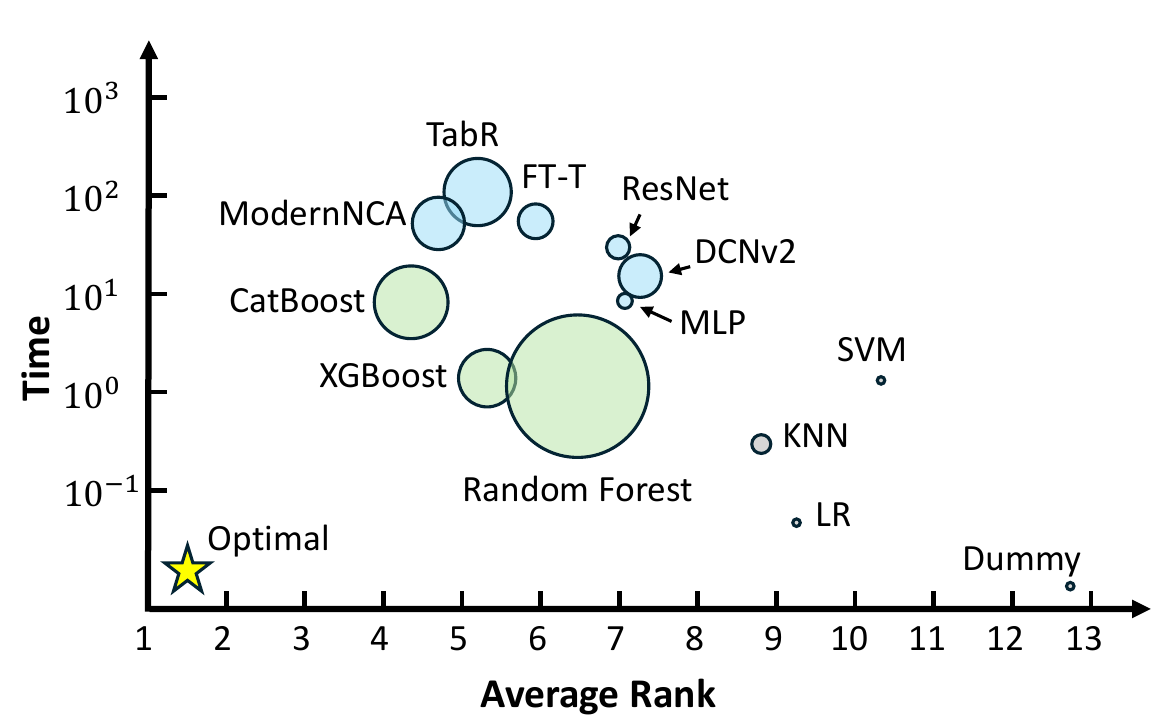}
    \centering
    {\small \mbox{(d) {All Tasks}}}
    \end{minipage}
  \caption{Performance-Efficiency-Size comparison of representative tabular methods on our toolbox for (a) binary classification, (b) multi-class classification, (c) regression tasks, and (d) all task types. The performance is measured by the average rank of all methods (lower is better). 
  We also consider the \textbf{dummy} baseline, which outputs the label of the major class and the average labels for classification and regression tasks, respectively.}
  \vspace{-8mm}
  \label{fig:teaser}
\end{figure}

We provide comprehensive evaluations of classical and deep tabular methods based on our toolbox in a fair manner in~\autoref{fig:teaser}.
Three tabular prediction tasks, namely, binary classification, multi-class classification, and regression, are considered, and each subfigure represents a different task type. The datasets are available at~\href{https://drive.google.com/file/d/18RHGSA1nASbsF1KAHCqLJasYsZIBXJ8D/view?usp=sharing}{Google Drive}.

We use accuracy and RMSE as the metrics for classification and regression, respectively. To calibrate the metrics, we choose the average performance rank to compare all methods, where a lower rank indicates better performance, following~\cite{FriedmanRank}. 
Efficiency is calculated by the average training time in seconds, with lower values denoting better time efficiency. The model size is visually indicated by the radius of the circles, offering a quick glance at the trade-off between model complexity and performance.

From the comparison, we observe that CatBoost achieves the best average rank in most classification and regression tasks, consistent with findings in~\cite{McElfreshKVCRGW23}. Among all deep tabular methods, ModernNCA performs the best in most cases while maintaining an acceptable training cost.
These visualizations serve as an effective tool for quickly and fairly assessing the strengths and weaknesses of various tabular methods across different task types, enabling researchers and practitioners to make informed decisions when selecting suitable modeling techniques for their specific needs.

%% file: conclusion.tex
\section{Conclusion}
We introduce \thename, a machine learning toolbox for tabular data prediction tasks. \thename implements both classical and deep tabular methods and includes several modules, such as hyperparameter tuning and preprocessing capabilities, to optimize the learning efficiency and effectiveness on tabular datasets.
We also leverage \thename to compare recent deep tabular methods fairly across numerous datasets. The toolbox is designed to be user-friendly and accessible to practitioners across diverse fields, providing a unified interface that is adaptable for integration with newly designed methods.